\documentclass{article}
\usepackage{iclr2019_conference,times}
\iclrfinalcopy

\usepackage{amsmath,amsfonts,bm}

\def\eqref#1{equation~\ref{#1}}
\def\1{\bm{1}}

\DeclareMathAlphabet{\mathsfit}{\encodingdefault}{\sfdefault}{m}{sl}
\SetMathAlphabet{\mathsfit}{bold}{\encodingdefault}{\sfdefault}{bx}{n}

\usepackage{amsthm}
\usepackage[utf8]{inputenc} % allow utf-8 input
\usepackage[T1]{fontenc}    % use 8-bit T1 fonts
\usepackage{url}            % simple URL typesetting
\usepackage{booktabs}       % professional-quality tables
\usepackage{amsfonts}       % blackboard math symbols
\usepackage{nicefrac}       % compact symbols for 1/2, etc.
\usepackage{microtype}      % microtypography
\usepackage{booktabs}
\usepackage{graphicx}
\usepackage{subcaption}
\usepackage{amsthm}

\usepackage{tikz}
\usetikzlibrary{positioning}
\usepackage{amsmath}

\definecolor{shadeOne}{rgb}{0.9,0.95,0.95}
\definecolor{shadeTwo}{rgb}{0.99,0.99,0.99}

\usepackage[bordercolor=shadeOne, backgroundcolor=shadeTwo, linecolor=red, textwidth=0.45in, textsize=tiny]{todonotes}

\title{Dissecting Pruned Neural Networks}

\author{Jonathan Frankle\\MIT CSAIL\\\texttt{jfrankle@csail.mit.edu} \And David Bau\\MIT CSAIL\\\texttt{davidbau@csail.mit.edu}}

\begin{document}

\maketitle

\begin{abstract}
Pruning is a standard technique for removing unnecessary structure from a neural network to reduce its storage footprint, computational demands, or energy consumption.
Pruning can reduce the parameter-counts of many state-of-the-art neural networks by an order of magnitude without compromising accuracy, meaning these networks contain a vast amount of unnecessary structure.

\vspace{.5em}
In this paper, we study the relationship between pruning and interpretability. Namely, we consider the effect of removing unnecessary structure on the number of hidden units that learn disentangled representations of human-recognizable concepts as identified by network dissection.
We aim to evaluate how the interpretability of pruned neural networks changes as they are compressed.

\vspace{.5em}
We find that pruning has no detrimental effect on this measure of interpretability until so few parameters remain that accuracy beings to drop.
Resnet-50 models trained on ImageNet maintain the same number of interpretable concepts and units until more than 90\% of parameters have been pruned.
\end{abstract}

\section{Introduction}
\label{sec:intro}

Neural network {pruning}~(e.g., \citet{brain-damage, han-pruning, pruning-filters}) is a standard set of techniques for removing unnecessary structure from networks in order to reduce storage requirements, improve computational performance, or diminish energy demands.
In practice, techniques for pruning individual connections from neural networks can reduce parameter-counts of state-of-the-art models by an order of magnitude \citep{han-pruning, gale-pruning} without reducing accuracy.
In other words, only a small portion of the model is necessary to represent the function that it eventually learns, meaning that---at the end of training---the vast majority of parameters are superfluous.
In this paper, we seek to understand the relationship between these superfluous parameters and the interpretability of the underlying model. 
To do so, we study the effect of pruning a neural network on its interpretability. We consider three possible hypotheses about this relationship:

\textit{Hypothesis A: No relationship.} Pruning does not substantially alter the interpretability of a neural network model (until the model has been pruned to the extent that it loses accuracy).

\textit{Hypothesis B: Pruning improves interpretability.} Unnecessary parameters only obscure the underlying, simpler function learned by the network. By removing unnecessary parameters, we focus attention on the most important components of the neural network, thereby improving interpretability.

\textit{Hypothesis C: Pruning reduces interpretability.} A large neural network has the capacity to represent many human-recognizable concepts in a detectable fashion. As the network loses parameters, it must learn compressed representations that obscure these concepts, reducing interpretability.

\paragraph{Interpretability methodology.}
We measure the interpretability of pruned neural networks using the \emph{network dissection} technique \citep{netdissect}.
Network dissection aims to identify convolutional units that recognize particular human-interpretable concepts.
It does so by measuring the extent to which each unit serves as binary segmenter for that concept on a series of input images.
The particular images considered are from a dataset called Broden assembled by \citeauthor{netdissect}; this dataset contains pixel-level labels for a wide range of hierarchical concepts, including colors, textures, objects, and scenes.
For each image in Broden, network dissection computes a convolutional unit's activation map, interpolates to expand it to the size of the input image, and segments the image based on the pixels for which the unit has a high activation according to its typical distribution of activations.%
\footnote{A high activation is determined by a threshold ``$T_k$ such that $P(a_k > T_k) = 0.005$ over every spatial location of the activation map in the data set.'' \citep{netdissect}}
Network dissection then computes the size of the intersection of the mask pixels and pixels labeled for particular concepts and divides this quantity by the size of the union; the technique considers units for which this ratio is larger than 0.05 to be interpretable, having learned a disentangled representation of this concept.

\paragraph{Pruning methodology.}
The neural networks that we dissect are Resnet-50 \citep{resnet} models trained on the ImageNet \citep{imagenet} dataset.
We apply a sparse pruning technique, removing weights with the highest magnitudes at the end of training (as in \citet{han-pruning}, \citet{gale-pruning}, and \citet{lth}).
Doing so produces pruned networks that have fewer parameters but the same number of neurons, meaning these pruned networks retain the capability to contain as many interpretable neurons as the original network.

Immediately after pruning, a neural network's accuracy decreases because part of the model has been removed; pruned networks are typically \emph{fine-tuned} for a small number of training steps to recover accuracy.
We use the lottery ticket fine-tuning procedure \citep{lth}, where the weights of a network are reset back to their values at an iteration early in training.
\citeauthor{lth} show (and we confirm for our models)
that networks trained in this way can learn to match the accuracy of the original network.
We choose this fine-tuning approach to allow pruned networks to learn from an early stage of training, potentially learning different functions better adapted to the smaller model. The Resnet-50 networks for ImageNet studied in this paper were uncovered using a modified version of this technique described by \citet{lthas}.

\begin{figure}
\centering
\includegraphics[width=.5\textwidth]{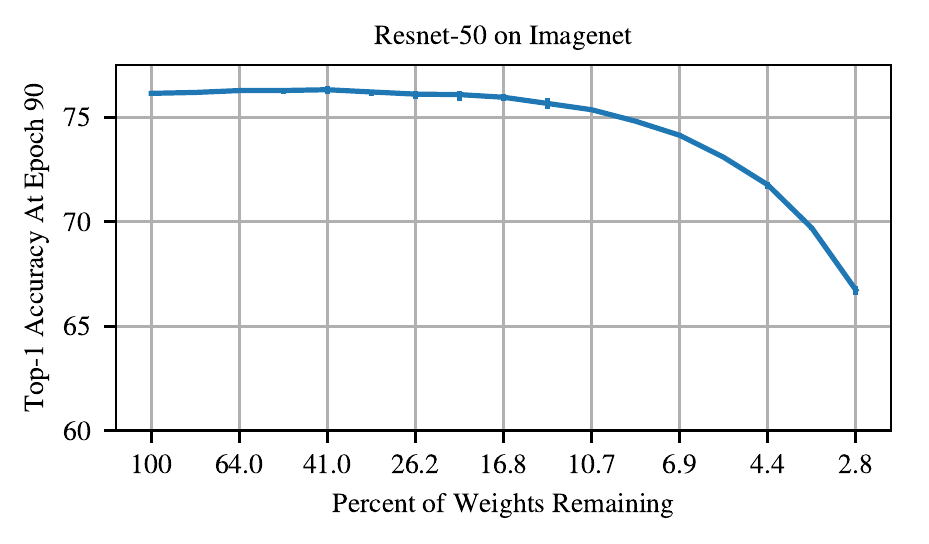}
\caption{The top-1 accuracy of Resnet-50 on ImageNet when pruned to the specified size.}
\label{fig:lth-imagenet}
\end{figure}

We prune Resnet-50 \emph{iteratively}, removing 20\% of weights, fine-tuning, and then pruning again. This process produces pruned networks at increments of 20\%,
making it possible to evaluate the effect of pruning on interpretability as a network is gradually reduced in size. Figure \ref{fig:lth-imagenet} shows the top-1 accuracy
of this network as a function of the number of parameters remaining. When 16.8\% of parameters or more remain, accuracy matches that of the original network. When 10\% of parameters remain, accuracy drops by a percentage point, followed by a steeper decline under further pruning.

\paragraph{Findings.}
We find that sparse pruning does not reduce the interpretability of Resnet-50 until so many parameters are pruned that accuracy declines, supporting Hypothesis A.
We conclude that the parameters that pruning considers to be superfluous for accuracy are also superfluous for interpretability.

\begin{figure}
\centering
\includegraphics[width=.5\textwidth]{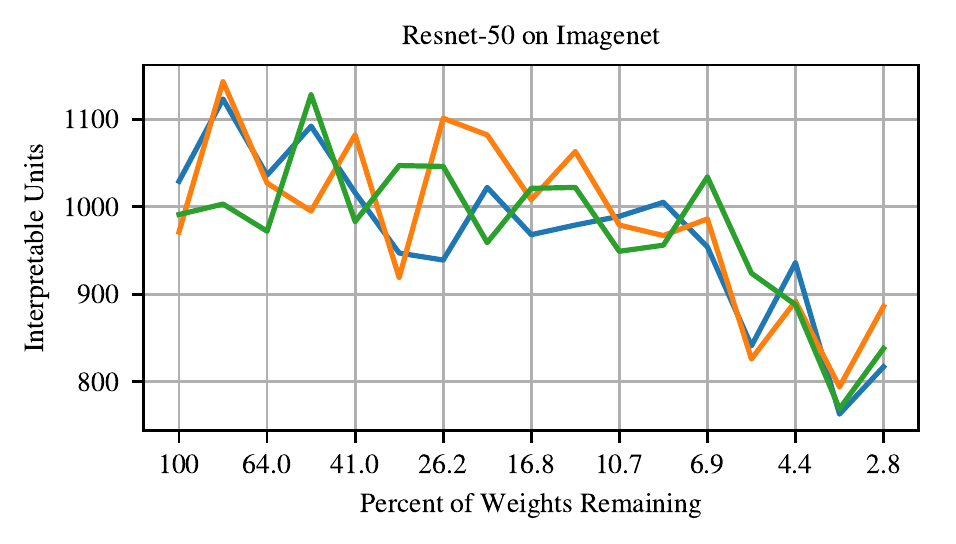}%
\includegraphics[width=.5\textwidth]{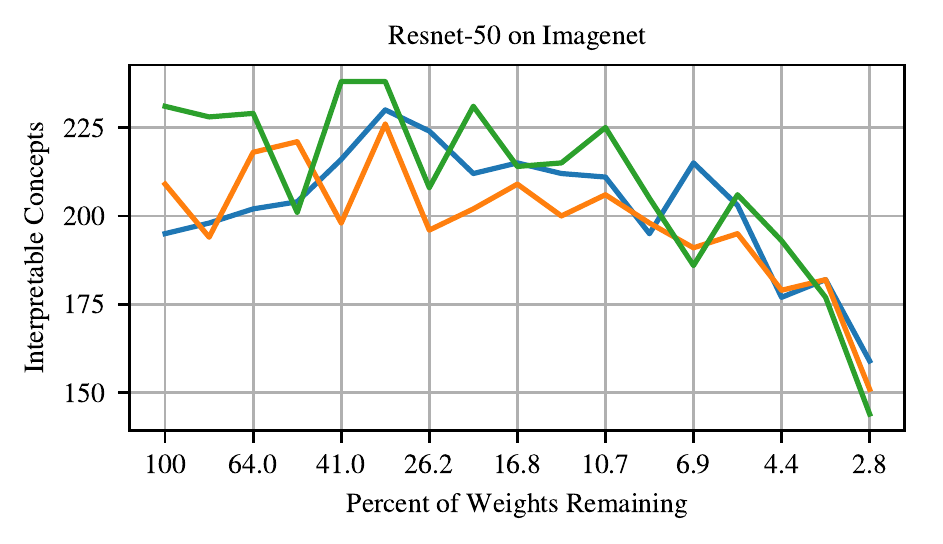}
\caption{Out of 2048 convolutional units in the fourth group of layers in Resnet-50, the number that learn disentangled concepts (left) and the overall number of concepts learned by any unit (right). Each line represents a separate trial of a model trained with a different initialization. Pruning does not reduce interpretability until it also reduces accuracy (compare to Figure \ref{fig:lth-imagenet}).}
\label{fig:lth-dissection}
\end{figure}

\section{Results}
\label{sec:results}

Network dissection considers both the number of units that learn disentangled concepts and the overall number of Broden concepts learned by any unit.
We study these quantities for the final four layers of Resnet-50, comprising 2048 units in total.
Based on the analysis of \citeauthor{netdissect}, we expect these layers to learn higher-level concepts like objects and scenes.

\paragraph{Interpretability.}
Figure \ref{fig:lth-dissection} plots the number of units that learn disentangled concepts (left) and the overall number of concepts learned (right). Each line represents a separate trained Resnet-50 model starting with a different random initialization.
All three trials show similar behavior: until 16.8\% of parameters remain, the network remains as interpretable as it was before it was pruned; after this point, interpretability begins to gradually decline.
This pattern indicates that sparse pruning has little relationship with interpretability (Hypothesis A)---interpretability barely suffers until more than 90\% of parameters have been pruned.
Instead, this pattern seems to follow the trend of network accuracy (Figure \ref{fig:lth-imagenet}).
Interpretability begins to decline at the same parameter-count that the network becomes less accurate as a product of over-pruning. 

\begin{figure}
\centering
\includegraphics[width=.3\textwidth]{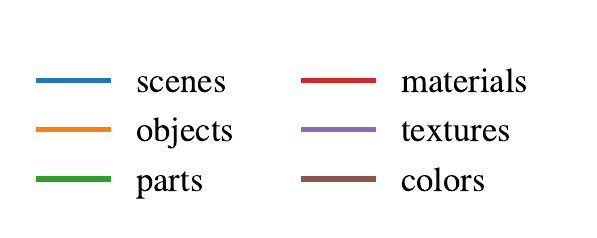}%
\includegraphics[width=.5\textwidth]{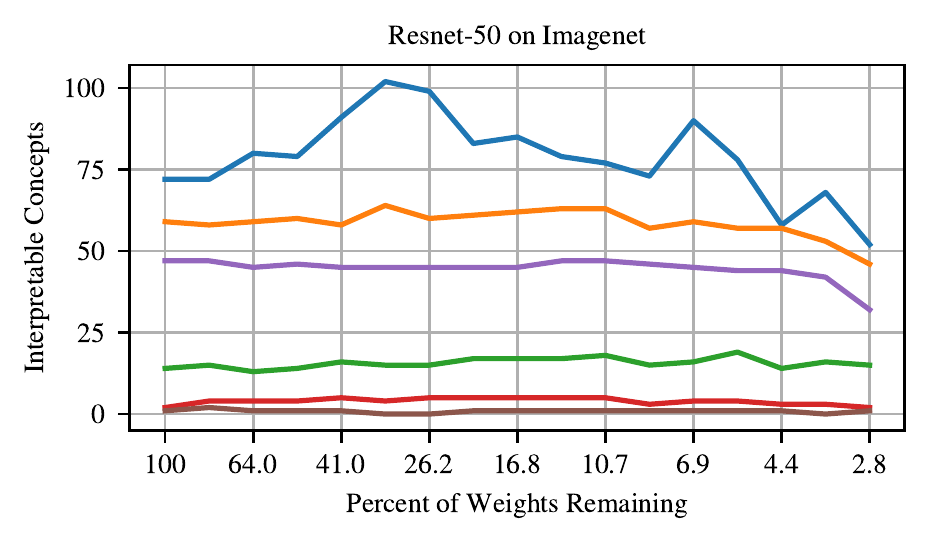}
\caption{The number of disentangled concepts learned by any unit. The categories are sorted into higher-level Broden categories representing the granularity of each concept. This graph breaks down these concepts for a single trial from Figures \ref{fig:lth-imagenet} and \ref{fig:lth-dissection}.}
\label{fig:lth-concepts}
\end{figure}

Figure \ref{fig:lth-concepts} separates a single trial from the right plot of Figure \ref{fig:lth-dissection} into a taxonomy of concepts according to their level of granularity.
Higher-level concepts like scenes and objects seem to be more volatile in the face of pruning.
Higher-level concepts are also more likely to disappear as interpretability and accuracy drop at extreme levels of pruning.
It is possible that the network's failure to learn as many disentangled, higher-level concepts diminishes its overall accuracy.

\begin{figure}
\centering
\includegraphics[width=.5\textwidth]{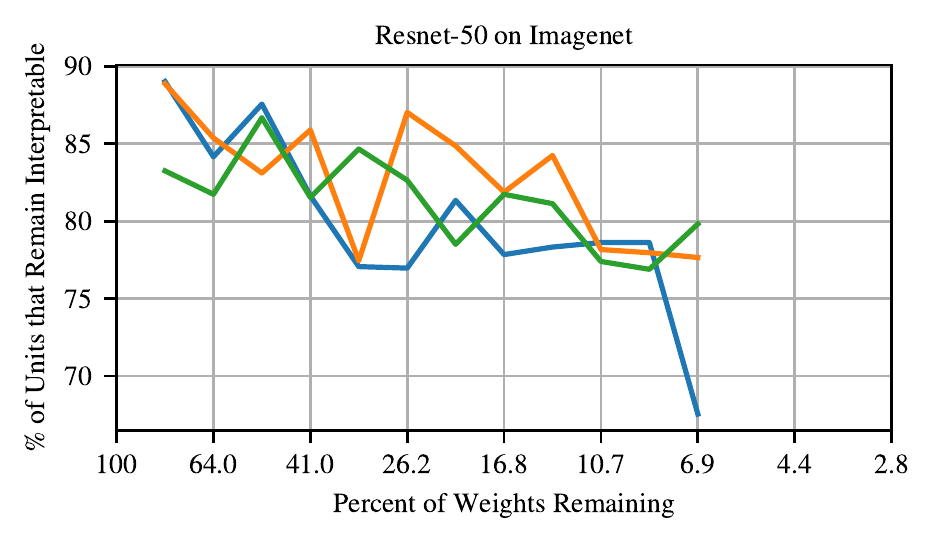}%
\includegraphics[width=.5\textwidth]{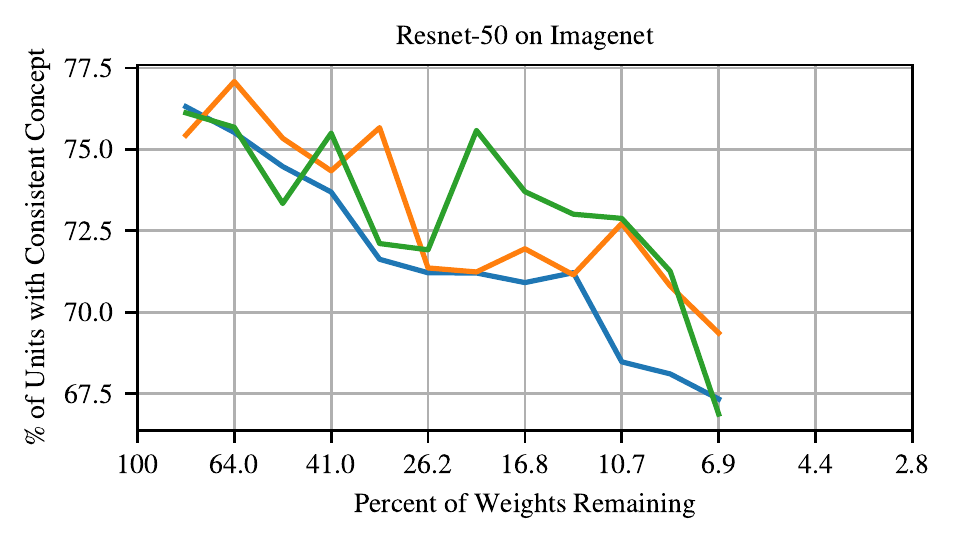}
\caption{Of the units that are interpretable in the original network, (left) the percent of interpretable units that were interpretable in the original network and (right) the percent of interpretable units that recognize the same concept as they did when they were in the original network. Each line represents a model trained with a different initialization.}
\label{fig:lth-consistency}
\end{figure}

\paragraph{Consistency.}
We use the lottery ticket procedure \citep{lth} to fine-tune after pruning, meaning that the pruned networks are re-trained nearly from initialization.
In comparison to other fine-tuning strategies, we believe this configuration offers pruned networks more leeway to learn new representations.
We are therefore interested in understanding the extent to which the same units are interpretable and learn to recognize the same concepts in the original and pruned networks.

The left graph in Figure \ref{fig:lth-consistency} shows the percentage of interpretable units in the original network that were also interpretable in the pruned network. Although this figure declines as the networks are pruned, nearly 80\% of the originally interpretable units remain interpretable even after 89\% of parameters have been pruned.
Of these units that were interpretable in both the original and pruned networks, the right graph in Figure \ref{fig:lth-consistency} explores the consistency of the concepts these units learn.
For each pruned network, it plots the percentage of interpretable units that recognize the same concept as they did when they were in the original network, considering only those units that were interpretable both in the original network and in the particular pruned network.
As the network is pruned, the fraction of such \emph{consistent} units declines.
However, it remains relatively high: about 70\% of such units learn to recognize the same concept even after 89\% of parameters are pruned.

\section{Discussion and Future Work}
\label{sec:future}

In this short paper, we only consider a sparse pruning technique that preserves the number of units in the network.
It is possible that, if entire convolutional filters were pruned as in \citep{pruning-filters}, a completely different set of behaviors might result. For example, the network might become less interpretable with pruning as it has less capacity with which to develop intermediate representations. Or alternatively, if---as \citet{rethinking-pruning} argue---pruned convolutional filters were never necessary to begin with, then the network would remain equally interpretable but with a higher percentage of interpretable units (convolutional filters are units with respect to network dissection).

It is also possible that another fine-tuning strategy might produce different results.
The lottery ticket strategy allows the network to retrain nearly from the start after each round of pruning, meaning that the network has the opportunity to learn entirely new representations.
In contrast, standard pruning techniques retain the trained weights of unpruned connections and fine-tune for a small number of iterations at a low learning rate, severely limiting the network's ability to learn new representations.
It would be interesting to compare the interpretability of the networks produced by each approach.
It is possible that lottery ticket fine-tuning makes it possible to learn new, disentangled representations for the smaller network size, or, alternatively, that limited fine-tuning more effectively sustains the interpretability of the unpruned networks.

For this workshop paper, we only consider Resnet-50. It would be valuable to study the extent to which the behavior we observe extends to other networks as in \citet{netdissect}.

\section{Conclusions}
\label{sec:conclusions}

We study the interpretability of pruned networks.
Specifically, we use network dissection \citep{netdissect} to examine the number of units that learn to recognize disentangled, human-identifiable concepts in networks whose weights have been removed using lottery ticket pruning \citep{lth, lthas}.
We find that this sparse pruning has no impact on the interpretability of the Resnet-50 model (as trained on ImageNet) until so many parameters are pruned that accuracy beings to decline.
We conclude that parameters considered unnecessary by magnitude pruning are also unnecessary to maintain the level of interpretability of the unpruned model.
However, pruning does not cause interpretability to improve either.

\bibliography{local}

\begin{thebibliography}{10}
\providecommand{\natexlab}[1]{#1}
\providecommand{\url}[1]{\texttt{#1}}
\expandafter\ifx\csname urlstyle\endcsname\relax
  \providecommand{\doi}[1]{doi: #1}\else
  \providecommand{\doi}{doi: \begingroup \urlstyle{rm}\Url}\fi

\bibitem[Bau et~al.(2017)Bau, Zhou, Khosla, Oliva, and Torralba]{netdissect}
David Bau, Bolei Zhou, Aditya Khosla, Aude Oliva, and Antonio Torralba.
\newblock Network dissection: Quantifying interpretability of deep visual
  representations.
\newblock In \emph{Proceedings of the IEEE Conference on Computer Vision and
  Pattern Recognition}, pp.\  6541--6549, 2017.

\bibitem[Frankle \& Carbin(2019)Frankle and Carbin]{lth}
Jonathan Frankle and Michael Carbin.
\newblock The lottery ticket hypothesis: Finding sparse, trainable neural
  networks.
\newblock In \emph{Int. Conf. Represent. Learn.}, 2019.

\bibitem[Frankle et~al.(2019)Frankle, Dziugaite, Roy, and Carbin]{lthas}
Jonathan Frankle, Gintare~Karolina Dziugaite, Daniel~M Roy, and Michael Carbin.
\newblock Stabilizing the lottery ticket hypothesis.
\newblock \emph{arXiv preprint arXiv:1903.01611}, 2019.

\bibitem[Gale et~al.(2019)Gale, Elsen, and Hooker]{gale-pruning}
Trevor Gale, Erich Elsen, and Sarah Hooker.
\newblock The state of sparsity in deep neural networks.
\newblock \emph{arXiv preprint arXiv:1902.09574}, 2019.

\bibitem[Han et~al.(2015)Han, Pool, Tran, and Dally]{han-pruning}
Song Han, Jeff Pool, John Tran, and William Dally.
\newblock Learning both weights and connections for efficient neural network.
\newblock In \emph{Advances in neural information processing systems}, pp.\
  1135--1143, 2015.

\bibitem[He et~al.(2016)He, Zhang, Ren, and Sun]{resnet}
Kaiming He, Xiangyu Zhang, Shaoqing Ren, and Jian Sun.
\newblock Deep residual learning for image recognition.
\newblock In \emph{Proceedings of the IEEE conference on computer vision and
  pattern recognition}, pp.\  770--778, 2016.

\bibitem[LeCun et~al.(1990)LeCun, Denker, and Solla]{brain-damage}
Yann LeCun, John~S Denker, and Sara~A Solla.
\newblock Optimal brain damage.
\newblock In \emph{Advances in neural information processing systems}, pp.\
  598--605, 1990.

\bibitem[Li et~al.(2016)Li, Kadav, Durdanovic, Samet, and
  Graf]{pruning-filters}
Hao Li, Asim Kadav, Igor Durdanovic, Hanan Samet, and Hans~Peter Graf.
\newblock Pruning filters for efficient convnets.
\newblock \emph{arXiv preprint arXiv:1608.08710}, 2016.

\bibitem[Liu et~al.(2019)Liu, Sun, Zhou, Huang, and
  Darrell]{rethinking-pruning}
Zhuang Liu, Mingjie Sun, Tinghui Zhou, Gao Huang, and Trevor Darrell.
\newblock Rethinking the value of network pruning.
\newblock In \emph{International Conference on Learning Representations}, 2019.
\newblock URL \url{https://openreview.net/forum?id=rJlnB3C5Ym}.

\bibitem[Russakovsky et~al.(2015)Russakovsky, Deng, Su, Krause, Satheesh, Ma,
  Huang, Karpathy, Khosla, Bernstein, et~al.]{imagenet}
Olga Russakovsky, Jia Deng, Hao Su, Jonathan Krause, Sanjeev Satheesh, Sean Ma,
  Zhiheng Huang, Andrej Karpathy, Aditya Khosla, Michael Bernstein, et~al.
\newblock Imagenet large scale visual recognition challenge.
\newblock \emph{International Journal of Computer Vision}, 115\penalty0
  (3):\penalty0 211--252, 2015.

\end{thebibliography}
\bibliographystyle{iclr2019_conference}

\end{document}